\documentclass{article} 
\usepackage{aditi}

\usepackage{microtype}
\usepackage{hyperref}
\usepackage{url}
\usepackage{booktabs}
\usepackage{style}
\usepackage{titlesec}
\usepackage{enumitem}
\usepackage{amsfonts}
\usepackage{nicefrac}
\usepackage{subcaption}
\usepackage{xcolor}
\usepackage{bm}
\usepackage{graphicx}
\usepackage{amsmath}
\usepackage{amssymb}
\usepackage{amsthm}
\usepackage{xspace}
\usepackage{adjustbox}
\usepackage{multirow}
\usepackage{pifont}
\usepackage{caption}
\usepackage{inconsolata}
\usepackage{wrapfig}
\usepackage[ruled,vlined]{algorithm2e}

\definecolor{darkblue}{rgb}{0, 0, 0.5}
\hypersetup{colorlinks=true, citecolor=darkblue, linkcolor=darkblue, urlcolor=darkblue}

\titlespacing*{\subsection}{0pt}{0.5\baselineskip}{0.25\baselineskip}
\titlespacing*{\section}{0pt}{0.5\baselineskip}{0.25\baselineskip}

\usepackage[utf8]{inputenc} 
\usepackage[T1]{fontenc}    
\usepackage{url}            
\usepackage{booktabs}       
\usepackage{amsfonts}       
\usepackage{nicefrac}       
\usepackage{microtype}      
\usepackage{xcolor}         
\usepackage{amsmath}
\usepackage{amssymb}
\usepackage{enumitem}
\usepackage{graphicx}
\usepackage{xspace}
\usepackage{subcaption}

\usepackage{booktabs}       
\usepackage{amsfonts}       
\usepackage{nicefrac}       
\usepackage{microtype}      
\usepackage{subcaption}
\usepackage{xcolor}         
\usepackage{bm}             
\usepackage{graphicx}       
\usepackage{amsmath}        
\usepackage{amssymb}        
\usepackage{amsthm}         
\usepackage{xspace}         
\usepackage{adjustbox}      
\usepackage{multirow}       
\usepackage{pifont}         
\usepackage{caption}        
\usepackage{inconsolata}                                    
\usepackage{wrapfig}        
\usepackage{tcolorbox}
\usepackage{enumitem}
\usepackage[ruled,vlined]{algorithm2e}  
\renewcommand{\cite}[1]{\citep{#1}}                         

\setcitestyle{authoryear,round,citesep={;},aysep={,},yysep={;}}

\newcommand{\passone}{pass@1\xspace}
\newcommand{\Passone}{Pass@1\xspace}
\newcommand{\passk}{pass@$k$\xspace}
\newcommand{\Passk}{Pass@$k$\xspace}

\title{Self-Trained Verification for Training- and \\ Test-Time Self-Improvement}

\setauthors{Chen Henry Wu, \authorsep Aditi Raghunathan}
\setaffils{Carnegie Mellon University}
\setemail{{\{chenwu2,aditirag\}@cs.cmu.edu}}
\setcode{https://github.com/ar-forum/stv}
\setwebsite{https://ar-forum.github.io/stv-webpage/}

\begin{document}

\maketitle

\begin{abstract}
Self-improvement at scale has been a longstanding goal for reasoning models, and there are two natural places to do it: at test time, through verification-refinement (V-R) loops; and at training time, through self-training methods. Both are gated by the same bottleneck: the verifier. V-R loops stall when verifier scores inflate while accuracy stagnates, and when feedback is too generic to act on; self-training fails similarly when bad self-generated data are added to training. Better verification would unlock both, but the capability we want to train, i.e., catching self-generated errors, lacks training signal. To address this challenge, we propose \textit{self-trained verification} (STV). Our key observation is that, while a model cannot catch these errors alone, it can when shown the reference solution. We turn this asymmetry into a supervision target and train the verifier to imitate a more informed version of itself. At test time, STV substantially improves V-R loops on hard problems, while alternatives (e.g., SFT, RL on verifier scores, and even meta-verifiers) do not. STV roughly doubles accuracy on hard math and lifts it $14\times$ on scientific reasoning tasks ($1.5\% \to 21\%$). At training time, we additionally train the generator using RL with STV verifier's feedback inside the V-R loop -- a procedure we call \emph{verifier-in-the-loop training} (ViL). Starting from an RL-converged generator, ViL yields a further $33\%$ gain in \passone. More notably, the generator's standalone \passone, with no verifier at test time, climbs $30\%$ relative past where standard RL had converged. Hence, the next frontier in reasoning on hard problems may lie in how we train for and with verification. 
\end{abstract}

\section{Introduction}
\label{sec:intro}

\begin{figure}[ht]
  \centering
  \includegraphics[width=0.95\linewidth]{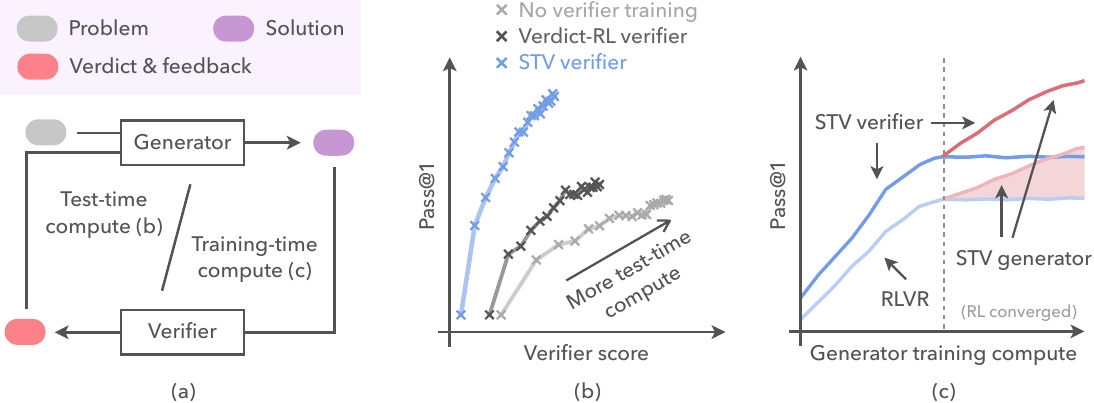}
  \caption{\textbf{(a)} We propose \textbf{self-trained verification} (STV), a scalable recipe for training verifiers that judge solutions and provide feedback for generators to refine. STV unlocks both self-improvement for test-time iterative refinement and training-time generator improvement. \textbf{(b)} STV scales much better with test-time compute than untrained or verdict-only-RL verifiers. \textbf{(c)} After standard RLVR converges, training the generator with \textbf{verifier feedback in the loop} (ViL) yields further \passone gain, along with a \emph{standalone} gain (shaded) visible without any verification at test time.}
  \label{fig:teaser}
\end{figure}

The long-term promise of reasoning models is self-improvement, where models can use their own attempts, mistakes, and feedback to solve harder problems. There are two natural places to do it. At test time, \textit{verifier-guided refinement}~\citep{Huang2025WinningGA,Shao2025DeepSeekMathV2TS} turns additional inference compute into better solutions, where a generator and verifier alternate, with the generator producing a solution, the verifier providing feedback to it, and the generator refining. This drives recent breakthroughs on the hardest reasoning tasks like IMO and frontier math~\citep{Feng2026TowardsAM}. At training time, \textit{self-training}~\citep{Zelikman2022STaRBR,Singh2023BeyondHD} recycles the generator's own, verified attempts into supervision. While recent frontier systems lean increasingly on verification, their gains are capped by verifier quality: shallow critiques~\citep{Huang2023LargeLM,Kamoi2024WhenCL,Stechly2024OnTS,Song2026ExpandingTC} and miscalibrated scores~\citep{Dang2026EscapingTC,Pan2024FeedbackLW} cause additional compute to amplify plausible but wrong reasoning.

\begin{tcolorbox}[colback=violet!8, colframe=violet!40, boxrule=0.5pt, arc=2mm]
\textit{In this work, we develop a scalable recipe for training verification and generation that unlocks self-improvement at both ends: test-time iterative refinement (by $14\times$ on hard reasoning tasks) and training-time generator improvement (by $30\%$ beyond the RLVR plateau). 
}
\end{tcolorbox}

Final-answer labels can teach a verifier to judge whether a solution is wrong, but not to find the flaw in a plausible but incorrect one -- a missed boundary case, an invalid reduction, or a lemma used outside its hypotheses. Existing systems sidestep this supervision gap with meta-verifiers trained on human feedback~\citep{Shao2025DeepSeekMathV2TS} or with external feedback such as a stronger model~\cite{Song2026ExpandingTC}. We instead exploit a simple asymmetry: \textbf{diagnosis is easier given a reference solution}. A model that struggles to find the flaw on its own can often locate it when shown a reference, because the task shifts from solving the problem to comparing a candidate against a known-good solution. Our pipeline, \textit{self-trained verification} (STV), turns this privileged setting into supervision. A reference-conditioned verifier acts as a teacher, and on-policy distillation trains an unconditioned verifier to match its feedback distribution. At test time, the resulting verifier runs without references.

STV makes verifier-guided refinement effective on problems out of reach for the base generator. On hard DAPO math problems~\citep{Yu2025DAPOAO} -- including a \textit{Hardest} split where Qwen3-8B has zero \passone -- an STV verifier roughly doubles final-round \passone over an untrained verifier, with gains compounding over 20+ refinement rounds. Standard alternatives (SFT on reference-conditioned feedback, RL on verdict accuracy) saturate quickly and yield much smaller gains (Figure~\ref{fig:teaser}, middle). The gains extend beyond math: on the hardest SciKnowEval problems, STV lifts \passone from 1.5\% without verification to 21.0\%. With enough verification compute, Qwen3-8B guided by STV outperforms the much larger Qwen3-32B generator, suggesting that trained verification can substitute for generator scale on hard reasoning problems.

Beyond test-time refinement, STV also enables a new way to training-time self-improvement: training the generator inside the V-R loop with the trained verifier providing feedback, a procedure we call \emph{verifier-in-the-loop training} (ViL). Starting from a generator already converged under standard reinforcement learning from verifiable reward (RLVR), we continue training it with feedback from a frozen STV verifier. Unlike standard self-training, the training signal here remains verifiable since the verifier's feedback only provides additional context that helps the generator to maximize that reward. ViL unlocks gains beyond the RLVR ceiling: final-round \passone improves by $33\%$ relative with the verifier present, and standalone \passone improves by $30\%$ relative even with no verifier at inference. However, the same compute spent extending standard RLVR yields no further gain (Figure~\ref{fig:teaser}, right, shaded).

In summary, STV improves both modes of verifier-driven self-improvement: at test time, it makes refinement scale; at training time, it breaks through the standard RLVR plateau, with gains that persist even with no verifier at inference. Our results point to a shift in where reasoning gains come from: from learning to verify -- and using stronger verifiers to train stronger generators. This suggests an iterative route to self-improvement, where better verifiers enable more reliable test-time refinement and richer generator training, while better generators produce harder attempts for future verifier training. Developing scalable ways to train verifiers without human-graded feedback may therefore be a key frontier for reasoning models that improve at both test time and training time.

\section{Problem Formulation}
\label{sec:setting}

We consider a reasoning problem $x$ with ground-truth answer $y^\star(x)$. A verification-refinement (V-R) pipeline couples a generator $G$ with a verifier $V$. These may be two separate models or the same model invoked with different prompts; we instantiate both cases in our experiments. At round $0$, the generator produces an initial solution $y_0 \sim G(\cdot \mid x)$. At each subsequent round $r \geq 1$, the verifier examines the latest solution and returns a verdict together with natural-language feedback,
\begin{equation*}
(v_r, f_r) \sim V(\cdot \mid x, y_{r-1}), \qquad v_r \in \{\text{accept}, \text{reject}\}.
\end{equation*}
If $v_r = \text{reject}$, the generator refines the solution based on the feedback: $y_r \sim G(\cdot \mid x, y_{r-1}, f_r)$. The loop ends at the first acceptance or after $R$ rounds.


\section{Self-Trained Verification}
\label{sec:method}

We propose \emph{self-trained verification} (STV), which trains this capability without human annotation: while a model cannot reliably generate feedback for the attempted solution from scratch, it \emph{can} locate logical gaps when shown the reference solution alongside the attempt, and this reference-conditioned distribution becomes a natural supervision target.

\label{sec:signal}

\begin{wrapfigure}[14]{r}{0.48\linewidth}
\vspace{-19pt}
\centering
\includegraphics[width=\linewidth]{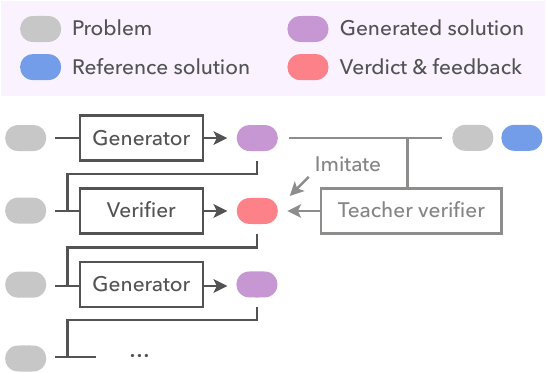}
\caption{Overview of self-trained verification.}
\label{fig:method}
\vspace{-6pt}
\end{wrapfigure}

At a high level, STV trains the verifier to imitate a more informed version of itself -- a teacher that is shown the reference solution and can therefore identify errors in $y_{r-1}$ more reliably. We parameterize the student verifier as $V_\theta(\cdot \mid x, y_{r-1})$ and the reference-conditioned teacher as
\begin{equation}
\label{eq:teacher-distribution}
V^\star(\cdot \mid x, y_{r-1}, y^\star(x)),
\end{equation}
both of which use the same model under different prompts.

We then align the student's output distribution with the teacher's. To do this, one can use either SFT or on-policy distillation (OPD). OPD matches the student's distribution $z = (v_r, f_r)$ to the teacher's:
\begin{equation}
\label{eq:opd}
\mathcal{L}_{\text{OPD}}(\theta) \;=\; \mathbb{E}_{(x, y_{r-1})} \, D_\alpha\!\left( V_\theta(\cdot \mid x, y_{r-1}) \;\big\|\; V^\star(\cdot \mid x, y_{r-1}, y^\star(x)) \right),
\end{equation}
where $(x, y_{r-1})$ is drawn from the generator $G$'s rollouts on training prompts, $V_\theta(\cdot \mid x, y_{r-1})$ and $V^\star(\cdot \mid x, y_{r-1}, y^\star(x))$ are full distributions over the response sequence $z$, and $D_\alpha$ is the $\alpha$-divergence with $\alpha = 0.5$ (Jensen-Shannon). On the other hand, SFT trains the student on $(v, f)$ pairs sampled from the teacher $V^\star$. The broader question of whether on-policy methods work better than SFT remains debated~\citep{Chu2025SFTMR,Matsutani2025RLSS}, and we make no general claim here. On the models and datasets we experimented with, we see that OPD improves the V-R performance while SFT fails (Section~\ref{sec:verifier_results_base}). We attribute this to the \emph{off-policy} nature of SFT: $V_\theta$ is trained on sequences $V^\star$ produces, but at test time it must generate its own, and once it drifts from $V^\star$ it encounters prefixes never seen during training. 

We also add a verdict-RL component to improve verdict accuracy, where we use RL with reward $\mathbb{I}[v_r = \mathbb{I}[y_{r-1} = y^\star(x)]]$. Putting everything together, the full STV objective is:
\begin{equation}
\label{eq:stv}
\mathcal{L}_{\text{STV}}(\theta) \;=\; \mathcal{L}_{\text{OPD}}(\theta) \;+\; \lambda \cdot \mathcal{L}_{\text{RL}}(\theta).
\end{equation}

\section{Verifier-in-the-Loop Training for Generator}
\label{sec:closing_loop}

Once we have a trained verifier $V_\theta$, the natural next step is to train the generator $G$ to better use $V_\theta$'s feedback inside the V-R loop. We call this procedure \emph{verifier-in-the-loop training} (ViL). We instantiate it as standard RL on the generator: each episode unrolls a multi-turn V-R rollout on a problem $x$ -- the generator produces $y_0 \sim G(\cdot \mid x)$, the (frozen) verifier produces $(v_1, f_1) \sim V_\theta(\cdot \mid x, y_0)$, the generator refines $y_1 \sim G(\cdot \mid x, y_0, f_1)$, and so on -- and the reward is the verifiable correctness of the final $y_r$ against $y^\star(x)$. Only $G$'s parameters are updated; $V_\theta$ stays frozen.

A natural expectation is that this training improves $G$'s performance at the V-R task at test time, since the generator now knows how to incorporate feedback. What is less obvious is whether the generator's behavior \emph{without} the loop should change at all. One might expect minimal transfer: $G$ could in principle achieve high reward by relying on $V_\theta$'s feedback to correct self-generated over later rounds, so there is no obvious pressure on $y_0$ itself to improve.

Our experiments (Section~\ref{sec:generator_training}) show otherwise. The expected V-R gain materializes -- final-round \passone climbs by $33\%$ relative -- but more surprisingly, $G$'s round-0 \passone also climbs by roughly $30\%$ relative, even when we start from a generator already converged under standard RL. We call this \emph{training-time self-improvement}: training $G$ inside the V-R loop teaches it skills that surface in its first attempt, without any verifier present at test time.

\section{Experiments}
\label{sec:results}

\subsection{Setup}

\begin{wrapfigure}[9]{r}{0.4\linewidth}
\vspace{-57pt}
\centering
\includegraphics[width=\linewidth]{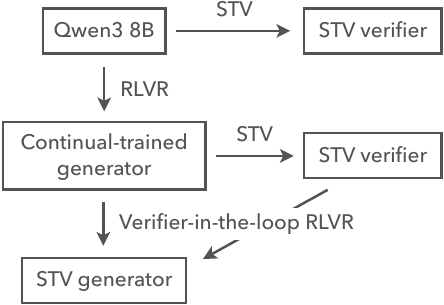}
\caption{Overview of our experiments.}
\label{fig:settings}
\vspace{-6pt}
\end{wrapfigure}
We use Qwen3-8B as the base model for both the generator and verifier in our main experiments unless specified. Given the asymmetry in output lengths, training a verifier is often more affordable than training a generator, so we consider two generators to simulate different training compute budgets: the base Qwen3-8B generator and a continual-trained generator trained with RLVR on the training set until convergence. An overview of experimental settings is shown in Figure~\ref{fig:settings}, which we will refer to in later sections.

Our training and test data are composed of hard math problems from DAPO~\citep{Yu2025DAPOAO}, split into two difficulty bins based on Qwen3-8B's \passone estimated with 32 rollouts: \textbf{Hard} ($0 <$ \passone $< 0.2$) and \textbf{Hardest} (\passone $= 0$). Training is performed on the joint of both bins. Since math data often share problems across sources, we embed all problems with OpenAI's \texttt{text-embedding-3-large} and remove test problems whose cosine similarity to any training problem exceeds $0.8$, leaving $\sim$150 problems per test bin. For each test problem, we run 32 independent V-R loops for up to 20 rounds. For simplicity, we use the same training problems for verifier and generator training, though this is not necessary and we leave data selection for verifier training to future work.

\begin{figure}[ht]
  \centering
  \includegraphics[width=0.95\linewidth]{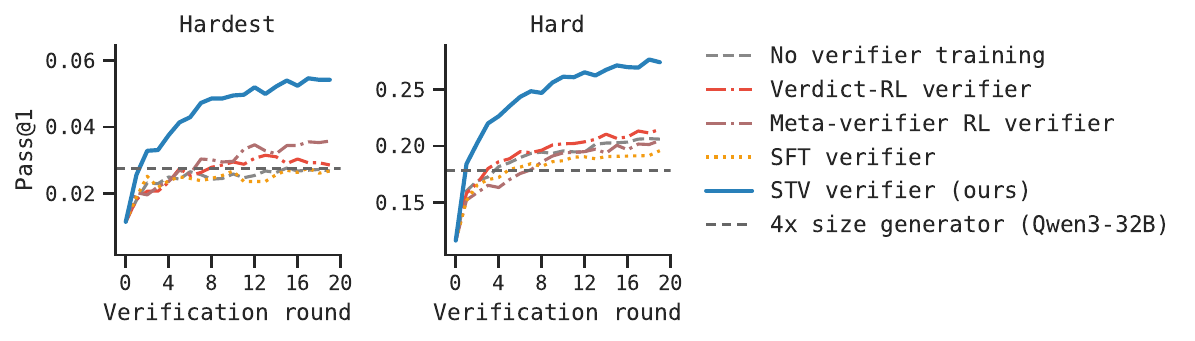}
  \caption{\Passone across verification rounds on the base generator.}
  \label{fig:verifier_base}
\end{figure}

\subsection{Verifier-Guided Refinement}
\label{sec:verifier_results_base}

\paragraph{Base generator.}
We first evaluate STV training on the base Qwen3-8B generator (Figure~\ref{fig:settings} top), holding the generator fixed and comparing different verifiers (Figure~\ref{fig:verifier_base}). Without any verifier training (\emph{No verifier training}), pipeline \passone stagnates quickly. Training the verifier with RL on verdict accuracy alone (\emph{Verdict-RL verifier}) helps only marginally. We also probe the meta-verifier approach by \citet{Shao2025DeepSeekMathV2TS} (\emph{Meta-verifier RL verifier}), which uses a separate model to grade the verifier's feedback as an additional RL reward signal; lacking access to their meta-verifier, we substitute GPT-5.2 as a proxy. This setup shows little effect. SFT on teacher-generated traces (\emph{SFT verifier}) shows no gains, consistent with the off-policy argument (Section~\ref{sec:signal}). In contrast, the STV verifier lifts final-round \passone by up to $2\times$ over the untrained-verifier pipeline  (\textit{p} value $<0.01$), with gains compounding over 20+ rounds. Notably, the STV-guided 8B at the final round ($5.5\%$ on Hardest, $27.4\%$ on Hard) outperforms the $4\times$ larger Qwen3-32B without verification ($2.7\%$ on Hardest, $17.8\%$ on Hard), which shows that a good verifier provides gains beyond what a much stronger generator can achieve. Appendix~\ref{app:verifier_examples} gives qualitative examples where, on identical generations, the STV verifier rejects flawed solutions that the untrained verifier accepts.

\begin{figure}[ht]
  \centering
  \includegraphics[width=0.92\linewidth]{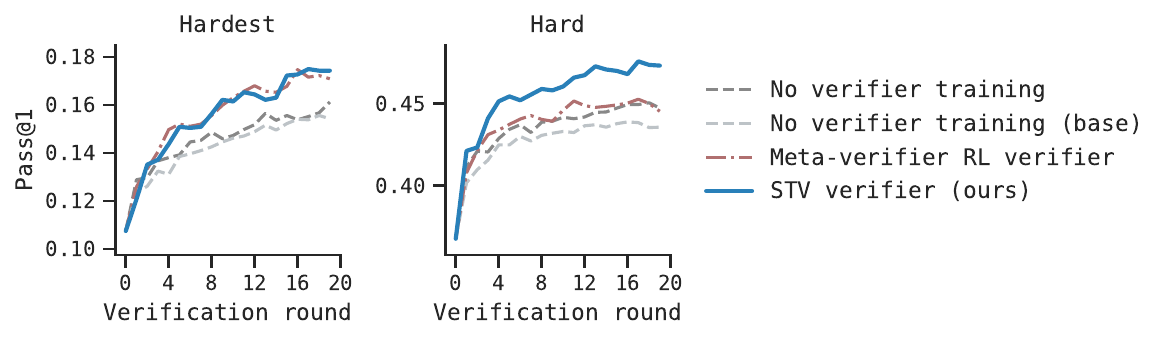}
  \caption{\Passone across verification rounds for the continual-trained generator.}
  \label{fig:verifier_rlgen}
\end{figure}

\begin{wraptable}[11]{r}{0.40\linewidth}
\vspace{-12pt}
\centering
\caption{Scientific reasoning results.}
\label{tab:sciknoweval}
\begin{tabular}{@{}lcc@{}}
\toprule
 & Hardest & Hard \\
\midrule
No verification & $\phantom{0}1.5$ & $11.5$ \\
\midrule
No verifier training & $\phantom{0}2.1$ & $11.4$ \\
STV verifier (ours) & $\mathbf{21.0}$ & $\mathbf{42.4}$ \\
\midrule
4$\times$ model size & $\phantom{0}8.9$ & $25.1$ \\
30$\times$ model size & $\phantom{0}8.0$ & $23.6$ \\
\bottomrule
\end{tabular}
\end{wraptable}
\paragraph{Scientific reasoning results.}
We then test STV on scientific reasoning tasks. We apply the same recipe to SciKnowEval~\citep{Feng2024SciKnowEvalEM}, a multi-domain science benchmark spanning chemistry, biology, physics, and materials science. We follow the math protocol to split problems by Qwen3-8B's \passone into \textbf{Hardest} (\passone $= 0$) and \textbf{Hard} ($0 <$ \passone $< 0.2$) and evaluate the V-R loop with Qwen3-8B as generator. Table~\ref{tab:sciknoweval} shows that untrained self-verification fails to improve over the no verification baseline, while STV lifts \passone by an order of magnitude on Hardest and nearly $4\times$ on Hard  (\textit{p} value $<0.01$). Going further, the STV-guided 8B even beats the $30\times$ larger Qwen3-235B-A22B on both bins ($21.0\%$ vs $8.0\%$ on Hardest, $42.4\%$ vs $23.6\%$ on Hard), as well as the $4\times$ larger Qwen3-32B ($8.9\%$ Hardest, $25.1\%$ Hard). This suggests that the bottleneck STV addresses is a domain-general capability.

\paragraph{Continual-trained generator.}
\label{sec:verifier_results_rlgen}
Would STV's gains compound on top of a generator already trained to be stronger, or would the generator's RL training absorb the benefit? We investigate this setting (Figure~\ref{fig:settings} middle row) in Figure~\ref{fig:verifier_rlgen}. The continual-trained generator (RLVR on the training set, to convergence) starts at higher \passone ($10.8\%$ and $37.2\%$) than the base model. We see that RLVR training of the generator already, but marginally, improves self-verification: the continual-trained generator's self-verification (\emph{No verifier training}) slightly outperforms using Qwen3-8B as verifier (\emph{No verifier training (base)}). In contrast, applying STV, initialized from the continual-trained generator, brings substantial gains over self-verification (\textit{p} value $<0.05$), with gains compounding over refinement iterations. This confirms that STV's gains are not absorbed by a stronger generator, and trained verification adds real lift on top of a converged generator.

\begin{figure}[ht]
  \centering
  \includegraphics[width=0.91\linewidth]{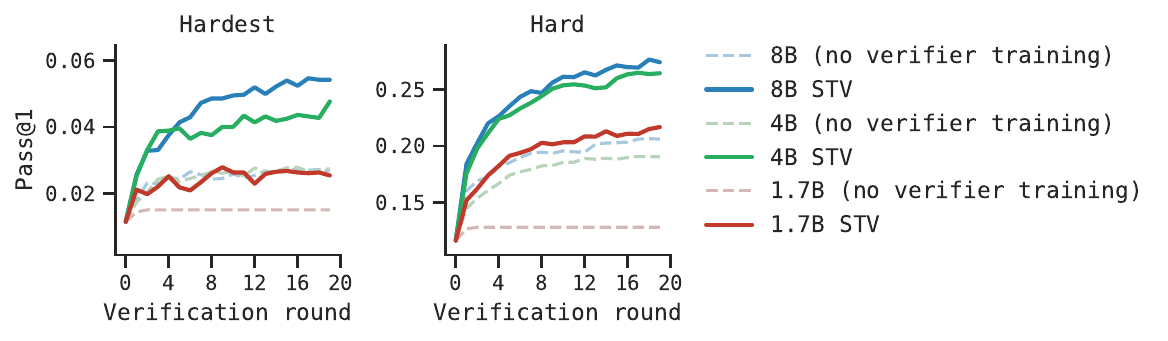}
  \caption{Training smaller verifiers to verify larger generators.}
  \label{fig:cross_scale}
\end{figure}

\subsection{Weak-to-Strong Verification}
\label{sec:cross_size}
Given STV's gains, can a smaller verifier match a larger one, which saves substantial test compute? Figure~\ref{fig:cross_scale} compares verifiers of different scales on the 8B generator. Without training, both the 4B and 1.7B verifiers perform no better than 8B self-verification. After STV, the 4B verifier becomes competitive with 8B STV (\passone of $26.4\%$ vs $27.4\%$), and the 1.7B verifier matches the untrained 8B verifier ($21.7\%$ vs $20.6\%$).

\subsection{Verifier-in-the-Loop (ViL) Generator Training}
\label{sec:generator_training}
So far in our experiments, the generator either remained fixed or was trained with pure RLVR. The strength of STV opens a second door: training the generator inside the V-R loop, which allows the model to provide guidance to enrich learning signal. We show that just \emph{one iteration} of ViL training (Section~\ref{sec:closing_loop}) already produces substantial training- and test-time self-improvement, producing the STV generator (Figure~\ref{fig:settings} bottom). Starting from the converged RL generator, we further train it with feedback from the frozen STV verifier; this is one round of training-time self-improvement.

\begin{figure}[ht]
  \centering
  \includegraphics[width=0.91\linewidth]{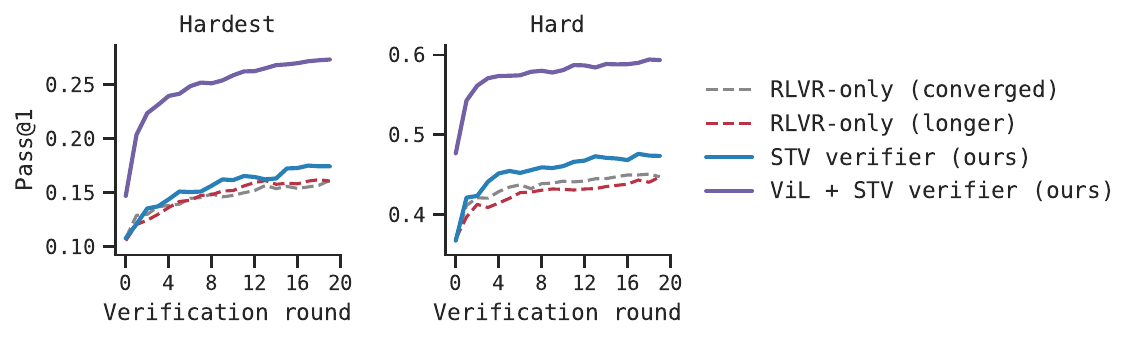}
  \caption{We train the generator to better incorporate verifier feedback, initialized with the already converged continual-trained generator, \emph{RLVR-only (converged)}. Surprisingly, such training shows a large benefit even at round 0 without any verification at test time. In contrast, spending the same additional compute on standard RLVR, i.e., \emph{RLVR-only (longer)}, has no gains.}
  \label{fig:closing_loop_pass1}
\end{figure}

Results are shown in Figure~\ref{fig:closing_loop_pass1}. Most strikingly, ViL training improves the generator \emph{even at round 0 -- before any verification at test time}. \Passone rises from $10.7\%$ to $14.7\%$ on Hardest ($+37\%$) and from $36.7\%$ to $47.7\%$ on Hard ($+30\%$), even though the generator was trained to improve \emph{after} receiving feedback. This suggests that learning from diagnostic feedback can instill more general reasoning capabilities. In contrast, continuing standard RL for the same number of additional steps (\emph{RLVR-only (longer)}) yields no gain, expected since the starting checkpoint is already converged under RLVR. With the STV verifier in the loop at test time, gains compound further: by the final round, \passone reaches $27.3\%$ on Hardest vs $16.1\%$ for \emph{RLVR-only (longer)} (\textit{p} value $<0.01$).

\paragraph{Ablation for the use of oracle.}
\label{sec:prefix_conditioning}
\begin{wraptable}[11]{r}{0.48\linewidth}
\vspace{-18pt}
\centering
\caption{Ablation for the use of oracle.}
\label{tab:prefix_conditioning}
\begin{tabular}{@{}lcc@{}}
\toprule
 & Round 0 & Round 20 \\
\midrule
\multicolumn{3}{@{}l}{\textit{Not using oracle solutions}} \\
\midrule
RLVR-only (converged) & $23.7$ & $30.4$ \\
ViL + self-verify (ours) & $29.8$ & $39.4$ \\
\midrule
\multicolumn{3}{@{}l}{\textit{Using oracle solutions}} \\
\midrule
Prefix-conditioning & $29.1$ & $38.5$ \\
ViL + STV verifier (ours) & $\mathbf{31.2}$ & $\mathbf{43.3}$ \\
\bottomrule
\end{tabular}
\end{wraptable}
\ Recall that STV training uses oracle solution, but what if we use the oracle elsewhere, or not at all? We first test an alternative for using oracle, \emph{prefix-conditioning}~\citep{Zhang2025BREADBR,Qu2026POPELT,Setlur2026ReuseYF}, which trains the generator to continue from the first $50\%$ of the reference solution. Table~\ref{tab:prefix_conditioning} shows that prefix-conditioning achieves a lower \passone than STV at both test-time round 0 (no verification) and round 20 (after V-R), indicating that gold prefixes do not substitute for diagnostic feedback. Second, we run ViL training with \emph{self-verification} instead of a trained verifier: the generator both produces feedback for itself in the V-R loop and is trained on its own corrections, never seeing a reference solution anywhere in the pipeline. Round-0 \passone slightly underperforms STV ($29.8\%$ vs $31.2\%$), suggesting that ViL drives training-time self-improvement even without a stronger verifier. At round 20, however, STV pulls ahead by about $4$ points ($43.3\%$ vs $39.4\%$), which confirms that the trained verifier's better feedback compounds across rounds even though both pipelines start from comparable standalone capability. Notably, ViL with self-verify (without oracle) even outperforms prefix-conditioning which require using oracle.

\subsection{Why STV Works}
\label{sec:why_stv_works}

To understand why STV improves test-time self-improvement, we decompose the contribution of STV into more calibrated verdict and higher feedback quality. 

\begin{figure}[ht]
  \centering
  \includegraphics[width=0.87\linewidth]{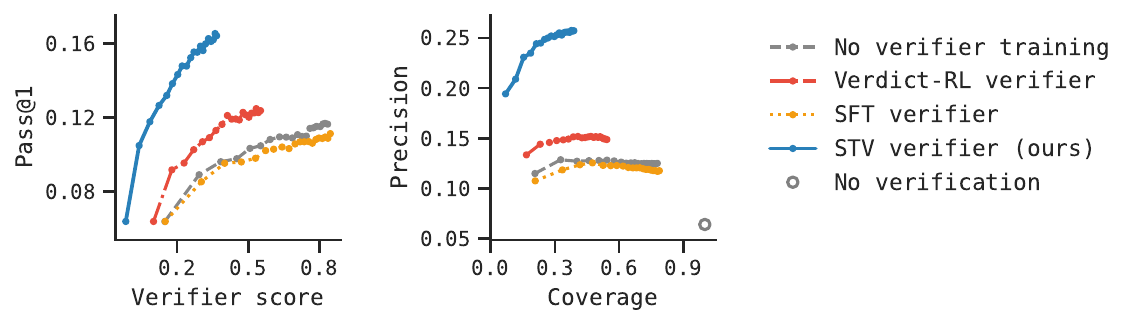}
  \caption{\textbf{Left}: Pass@1 vs verifier score over verification rounds (reward hacking diagnostic). \textbf{Right}: precision-coverage frontier of test-time scaling. Both panels pool Hardest and Hard problems.}
  \label{fig:reliable_scaling}
\end{figure}

\paragraph{Calibrated test-time scaling.}
\label{sec:precision_coverage}
We show that STV mitigates the miscalibrated verifier score problem. Without training, verifier scores increase over rounds while accuracy stagnates (Figure~\ref{fig:reliable_scaling} left). STV verifiers are better calibrated. Figure~\ref{fig:reliable_scaling} (right) shows the precision-coverage curves, where coverage is the fraction of solutions accepted by the round $t$ and precision is the fraction of acceptances that are correct. STV shows a ${\sim}3$--$5\times$ higher precision at the same level of coverage, and crucially, precision \emph{increases} with coverage, which means that each test-time verification round makes accepted solutions more likely correct, unlike untrained verifiers where more rounds provide no reliability gain.

\begin{figure}[ht]
  \centering
  \includegraphics[width=0.97\linewidth]{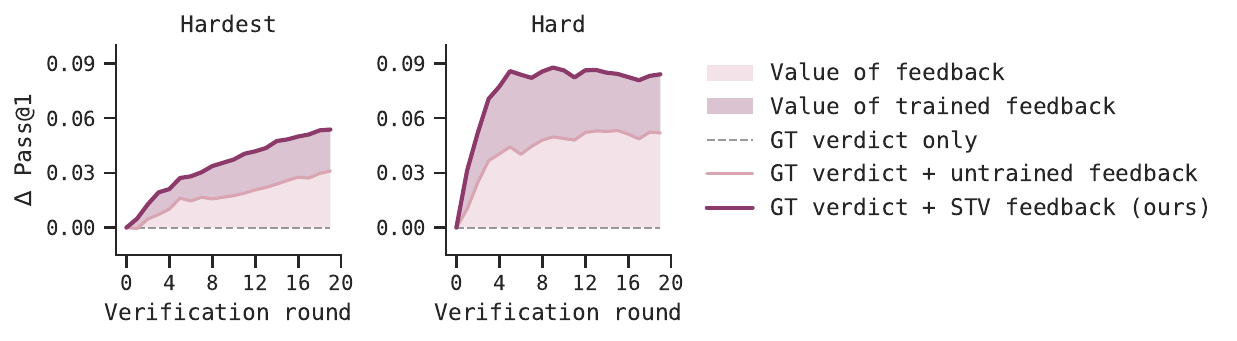}
  \caption{Isolating the contribution of trained feedback using ground-truth verdict.}
  \label{fig:oracle}
\end{figure}

\paragraph{Value of trained feedback.}
\label{sec:oracle}
To isolate the contribution of feedback from verifier score accuracy, we replace the verifier's verdicts with ground-truth correctness labels while varying the feedback. Figure~\ref{fig:oracle} plots $\Delta$\passone relative to a \emph{GT verdict only} baseline, which uses ground-truth verdicts with uninformative feedback (``Your solution appears to be incorrect.''). This baseline is shown as the dashed zero line. Adding untrained feedback from Qwen3-8B (\emph{GT verdict + untrained feedback}) gives a $+5.2\%$ gain on Hard at the final round, which confirms that feedback provides value beyond pure verdict. STV feedback (\emph{GT verdict + STV feedback}) adds a further $+3.2\%$ on top of it, showing that STV improves feedback quality independent of verdict accuracy.

\subsection{How STV Shapes the Output Distribution}
\label{sec:bon_vs_vr}

Beyond why the gains exist, we look at how V-R reshapes the output distribution. Our claim is that, unlike Best-of-$N$ (BoN) or resampling, V-R with STV improves the performance beyond sharpening the base distribution. This is because of the structured exploration enabled by actionable verifier feedback. We will verify this claim by exploring \passk and comparing with BoN directly. 

\begin{figure}[ht]
  \centering
  \begin{subfigure}[t]{0.495\linewidth}
    \centering
    \includegraphics[width=\linewidth]{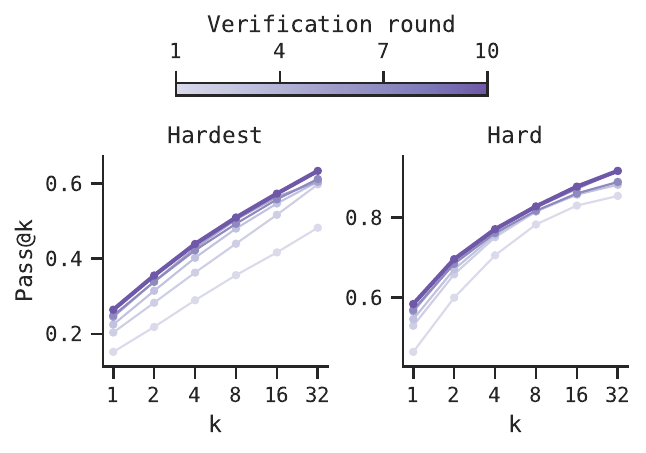}
    \caption{\Passk vs $k$ per verification round.}
    \label{fig:closing_loop_passk}
  \end{subfigure}
  \hfill
  \begin{subfigure}[t]{0.495\linewidth}
    \centering
    \includegraphics[width=\linewidth]{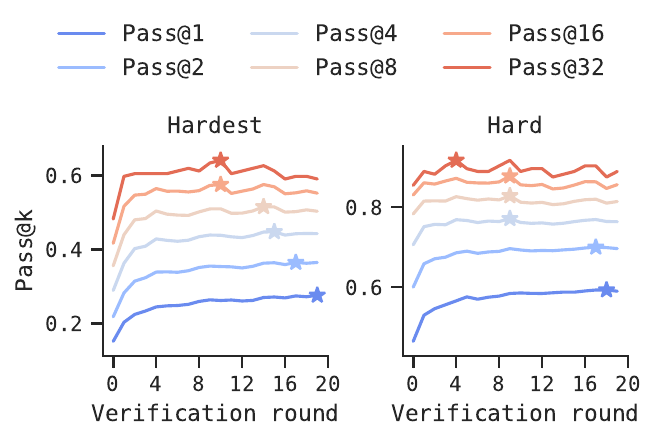}
    \caption{\Passk vs verification round per $k$.}
    \label{fig:sweetspot}
  \end{subfigure}
  \caption{Verification does not suppress diversity, but there is a sweet spot. \textbf{(a)} \Passk generally improves in the first 10 rounds. \textbf{(b)} At higher $k$, \passk eventually peaks at a certain but non-zero round; harder problems require more verification rounds to reach their peaks than easier ones.}
  \label{fig:diversity}
\end{figure}

\paragraph{Verification does not suppress diversity.}
\label{sec:diversity}
A natural concern is whether the gains of verification come at the cost of diversity, i.e., verification could be simply sharpening to the best mode. With our strongest setting, \emph{ViL + STV verifier}, this is not the case: \passk generally improves in the first 10 rounds of self-improvement (Figure~\ref{fig:closing_loop_passk}). However, there is a sweet spot for the number of verification rounds: \passk eventually peaks at a certain but non-zero round (Figure~\ref{fig:sweetspot}). Interestingly, harder problems often require more verification rounds to reach their \passk peaks.

\begin{figure}[ht]
  \centering
  \includegraphics[width=0.98\linewidth]{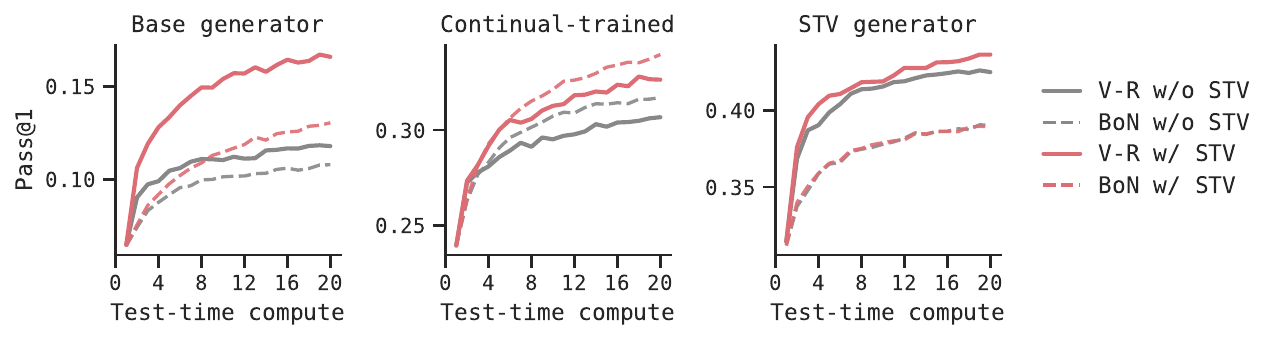}
  \caption{Refinement (V-R) vs resampling (BoN) at matched compute, for three generators considered in this paper. All panels pool Hardest and Hard problems.}
  \label{fig:bon_vs_vr}
\end{figure}

\paragraph{Refinement vs resampling.}
To test the sharpen-vs-reshape distinction, we compare refinement to BoN at matched compute. Given $N$ independent samples, BoN selects one the verifier accepts (or a uniformly random one if none is accepted); $r$ rounds of refinement matches BoN at $N = r+1$. In each setting, we also compare the STV verifier with the base verifier, which isolates the effect of verifier capability. Figure~\ref{fig:bon_vs_vr} shows that refinement dominates resampling for the base and STV generators -- evidence that reshaping the distribution beats merely sharpening it. The lone exception is the continual-trained generator: it was trained solely for round-0 accuracy and never learned to use feedback, so reshaping isn't available to it and only sharpening remains. Notably, STV verifier helps \textit{both} refinement and resampling in every case, which shows that STV verifier is a generally better verifier across different test-time compute scaling methods.

\section{Related Work}
\label{sec:related}

\textbf{Test-time compute scaling. \ \ }
Scaling test-time compute~\citep{ElKishky2024OpenAIOS,DeepSeekAI2025DeepSeekR1IR,Snell2024ScalingLT,Wu2024InferenceSL} -- through self-consistency~\citep{Wang2022SelfConsistencyIC}, best-of-$N$~\citep{Brown2024LargeLM}, extending unstructured chains of thought~\citep{Setlur2025e3LT}, or multi-turn verification~\citep{Shao2025DeepSeekMathV2TS} -- consistently improves reasoning performance. Our work contributes to this line by targeting structured multi-turn verification, and making its non-verifiable feedback trainable.

\textbf{Self-correction and self-improvement. \ \ }
Early work on self-improvement conditions the generator on its own previous output without any explicit verification step; Self-Refine~\citep{Madaan2023SelfRefineIR} and Reflexion~\citep{Shinn2023ReflexionLA} are canonical examples. \citet{Huang2023LargeLM} and \citet{Kamoi2024WhenCL} show that such self-correction struggles on reasoning tasks: without external feedback, LLMs rarely fix their own errors. Follow-up work trains the generator to self-correct via SFT~\citep{Qu2024RecursiveIT} or multi-turn RL~\citep{Kumar2024TrainingLM,Jiang2025PAGMR}. Broader self-improvement methods like STaR~\citep{Zelikman2022STaRBR}, ReST-EM~\citep{Singh2023BeyondHD}, and V-STaR~\citep{Hosseini2024VSTaRTV} generate training data from the model itself for continuous improvement. Our work bridges these two lines: running the V-R loop \emph{during training}, not only at test time, yields \emph{training-time self-improvement}, where the generator's standalone \passone climbs beyond the standard-RL ceiling.

\textbf{Verification-refinement pipelines. \ \ } V-R pipelines introduce an explicit verification step for structured exploration, which has led to recent breakthroughs in the hardest reasoning tasks such as IMO and frontier mathematics~\citep{Huang2025WinningGA,Feng2026TowardsAM,Shao2025DeepSeekMathV2TS}. A line of work trains the verifier explicitly: \citet{Shao2025DeepSeekMathV2TS} optimize verifiers by training a meta-verifier with RL based on human annotation of feedback quality; \citet{Kirchner2024ProverVerifierGI} does not target V-R pipelines directly but trains a verifier to improve the legibility of generator outputs. RL from text feedback~\cite{Song2026ExpandingTC} uses an external feedback provider during training and train generators to internalize and mimic the feedback. We focus on improving feedback without human annotations or supervision from a stronger model.

\textbf{Verifier and reward hacking. \ \ }
Feedback loops are prone to in-context reward hacking. For example, an LM agent that tries to maximize Twitter engagement may make posts more controversial \citep{Pan2024FeedbackLW}. \citet{Dang2026EscapingTC} show in the math setting that refinement can converge to plausible-but-wrong solutions, and propose to isolate potential errors from the context window. We show that STV addresses in-context reward hacking by improving verifier training.

\textbf{(Self-)distillation and privileged information. \ \ }
Knowledge distillation~\citep{Hinton2015DistillingTK} transfers a teacher's behavior to a student by matching distributions. More recently, on-policy distillation~\citep{Agarwal2023OnPolicyDO, Gu2023MiniLLMOD} samples prefix tokens from the student so the distillation target matches deployment. A natural way to get a teacher without relying on a stronger model is to grant it \emph{privileged information} such as reference solutions. When teacher and student share a base model, this becomes \emph{self-distillation}, studied recently in various forms~\citep{Zhao2026SelfDistilledRO,Hubotter2026ReinforcementLV,Shenfeld2026SelfDistillationEC,Song2026ExpandingTC,Penaloza2026PrivilegedID,Yang2026InTSI}. Among these, \citet{Yang2026InTSI} used reference solution to help truncate and intervene generated solutions to create finetuning data for generators. 
A closely related line splices partial reference solutions into the generator's context during RL~\citep{Zhang2025BREADBR, Qu2026POPELT, Setlur2026ReuseYF}. STV instantiates the privileged-information for a previously intractable target -- non-verifiable feedback quality for training verifiers. Moreover, we show that ViL training transfers the gains to the generator, providing a new axis of conditioning (i.e., self-generated feedback) for generator training.

\textbf{Process reward models. \ \ } 
Our work is also conceptually related to process reward models~\citep[PRMs;][]{Lightman2023LetsVS}. Unlike outcome reward models~\citep[ORMs;][]{Cobbe2021TrainingVT} that only assign a scalar reward for the entire rollout, PRMs enable dense credit assignment along the trajectory. Our verifier shares this view that verification signal should not be sparse, but goes further by producing diagnostic feedback that actively coaches the generator. \citet{Shao2025DeepSeekMathV2TS} is the closest: it trains a verifier using a meta-verifier with human annotations. STV introduces a dense signal for feedback quality \emph{without} such annotations.

\section{Conclusion, Limitations, and Future Work}
\label{sec:conclusion}

We proposed \emph{self-trained verification} (STV), a way to train the verifier to catch flaws in plausible but incorrect solutions without human annotation for feedback quality. STV turns this challenge into a distillation target: when conditioned on the reference solution, the model becomes its own teacher for feedback quality. On math and scientific reasoning benchmarks, we show that STV substantially improves over baselines for test-time self-improvement. We then show that \emph{verifier-in-the-loop training} (ViL) -- training the generator inside the V-R loop with STV verifier feedback -- provides a new training-time self-improvement direction, which breaks through the standard RLVR plateau.

Whether STV's approach extends to other capabilities of similar structure, i.e., non-verifiable from scratch but easier when the model is given an appropriate reference, is an open question. Within the V-R setting itself, other self-supervision signals may also be rich -- model uncertainty, multiple references, environment feedback, or training the verifier end-to-end against downstream generator improvement. Other open questions include scaling to larger models, generalization beyond math (code and open-ended reasoning), and the compute-optimal split between generator training, verifier training, and test-time rounds. More broadly, learning to verify -- and developing scalable ways to do so without human-graded feedback -- may be a key frontier for reasoning models that self-improve at both training and test time.

\section*{Acknowledgments} 
We wish to thank Jacob Springer, 
Christina Baek,
Jeremy Avigad, Sean Welleck, 
Marijn Heule, Yuda Song, Ziqian Zhong, 
and Daniel Fried for their discussion and feedback. 
This research is partially supported by
the DARPA expMath program through the DARPA CMO contract number HR0011262E028. We gratefully acknowledge support from Schmidt Sciences, NSF, Apple, Open Philanthropy, Google.
We would like to thank CMU FLAME Cluster for providing generous compute. 

\bibliography{iclr2026_conference}
\bibliographystyle{plainnat}

\clearpage
\appendix

\section{Verifier Feedback Examples}
\label{app:verifier_examples}

To qualitatively illustrate how STV training improves verification, we compare two verifiers on an \emph{identical} set of generations. We contrast the untrained Qwen3-8B verifier (\emph{No verifier training}, which verifies its own outputs) with the \emph{STV verifier}. In all examples the student's final answer is genuinely incorrect (a real reasoning error, not a formatting artifact). All feedback is shown verbatim. ``Predicted verdict'' denotes the verdict the verifier assigns to the student solution (\texttt{CORRECT} = accept, \texttt{INCORRECT} = reject), not whether that verdict is itself correct.

\newcommand{\vaccept}{Predicted verdict: \texttt{CORRECT}}
\newcommand{\vreject}{Predicted verdict: \texttt{INCORRECT}}

\subsection{The untrained verifier accepts flawed solutions}

Here the untrained verifier accepts a flawed solution, whereas the STV verifier rejects it and pinpoints the underlying mistake.

\textbf{Example 1 (counting and divisibility). \ \ }
\emph{Problem.} For how many triples of positive integers $(a,b,c)$ with $1 \le a,b,c \le 5$ is the quantity $(a+b)(a+c)(b+c)$ not divisible by $4$?
\begin{itemize}[leftmargin=1.4em,topsep=2pt,itemsep=1pt]
  \item \emph{Shared generator solution.} Reports a final answer of $14$ (ground truth: $48$).
  \item \emph{No verifier training.} \vaccept{} -- accepts the flawed solution.
  \item \emph{STV verifier.} \vreject{}: ``The student's final answer of 14 is incorrect. The reasoning about when the product is not divisible by 4 is flawed, particularly in how the parity conditions and divisibility by 4 are handled. The student's approach to counting valid cases is incomplete and leads to an incorrect result.''
\end{itemize}

\textbf{Example 2 (limit of a recurrence). \ \ }
\emph{Problem.} A sequence $\{a_n\}$ is defined such that $a_i = i$ for $i = 1, 2, \ldots, 2020$, and for $i > 2020$, $a_i$ is the average of the previous $2020$ terms. What is the largest integer less than or equal to $\lim_{n\to\infty} a_n$?
\begin{itemize}[leftmargin=1.4em,topsep=2pt,itemsep=1pt]
  \item \emph{Shared generator solution.} Equates the limit with the simple average of the first $2020$ terms, $\lfloor 2021/2 \rfloor = 1010$ (ground truth: $1347$).
  \item \emph{No verifier training.} \vaccept{} -- accepts the flawed solution.
  \item \emph{STV verifier.} \vreject{}: ``The student incorrectly assumes the limit is the average of the first 2020 terms, but fails to recognize that the sequence converges to a different value due to the recursive averaging process, which leads to a different limiting behavior.''
\end{itemize}

\textbf{Example 3 (unjustified symmetry assumption). \ \ }
\emph{Problem.} In $\triangle ABC$, $AB = 30$, $BC = 40$, and $CA = 50$. Squares $A_1A_2BC$, $B_1B_2AC$, and $C_1C_2AB$ are erected outside $\triangle ABC$, and the pairwise intersections of lines $A_1A_2$, $B_1B_2$, and $C_1C_2$ are $P$, $Q$, and $R$. Compute the length of the shortest altitude of $\triangle PQR$.
\begin{itemize}[leftmargin=1.4em,topsep=2pt,itemsep=1pt]
  \item \emph{Shared generator solution.} Assumes $\triangle PQR$ is equilateral without justification, yielding $50$ (ground truth: $124$).
  \item \emph{No verifier training.} \vaccept{} -- accepts the flawed solution.
  \item \emph{STV verifier.} \vreject{}: ``The student incorrectly assumed the outer Napoleon triangle is equilateral and used an incorrect formula for its side length, leading to an incorrect altitude calculation. The correct approach involves coordinate geometry and vector analysis, which the student did not apply.''
\end{itemize}

\textbf{Example 4 (misplaced geometric configuration). \ \ }
\emph{Problem.} Four lighthouses are at points $A$, $B$, $C$, $D$, with $AB = 5$, $BC = 12$, $AC = 13$ km. To an observer at $A$, the angles subtended by $\{B,D\}$ and by $\{C,D\}$ are equal; to an observer at $C$, the angles subtended by $\{A,B\}$ and by $\{D,B\}$ are equal. The distance $AD = \frac{p\sqrt{q}}{r}$ with $p,q,r$ relatively prime and $r$ square-free. Find $p+q+r$.
\begin{itemize}[leftmargin=1.4em,topsep=2pt,itemsep=1pt]
  \item \emph{Shared generator solution.} Places $D$ on the sides of $\triangle ABC$ and applies the angle-bisector theorem, yielding $15$ (ground truth: $96$).
  \item \emph{No verifier training.} \vaccept{} -- accepts the flawed solution.
  \item \emph{STV verifier.} \vreject{}: ``The student incorrectly assumed that point D lies on the sides of triangle ABC and applied the angle bisector theorem directly, leading to an incorrect result. The correct approach involves recognizing that D is not on the triangle sides and using coordinate geometry to find the correct distance.''
\end{itemize}

\subsection{Both verifiers reject, but the untrained verifier's feedback is incoherent}

A subtler failure appears when \emph{both} verifiers reject the solution. 

\textbf{Example 5 (counting and divisibility, revisited). \ \ }
\emph{Problem.} The same problem as Example~1: for how many triples $(a,b,c)$ with $1 \le a,b,c \le 5$ is $(a+b)(a+c)(b+c)$ not divisible by $4$? (Ground truth: $48$.) Here a different rollout reports $54$.
\begin{itemize}[leftmargin=1.4em,topsep=2pt,itemsep=1pt]
  \item \emph{No verifier training.} \vreject{}, yet the full feedback endorses the wrong answer: ``The student incorrectly assumes that the product is not divisible by 4 only when all three sums are odd, but this is not the only case. The product can also be odd if all three sums are odd, but the student's logic is correct in that scenario. However, the mistake lies in the final count: the number of favorable cases is not 54. The correct count should be 3 * 2 * 3 * 3 = 54, which matches the student's result. Therefore, the final answer is actually correct.''
  \item \emph{STV verifier.} \vreject{}: ``The student incorrectly assumed that the product is not divisible by 4 only when all three factors are odd, but failed to account for cases where one factor is divisible by 4 or two factors are even (but not both divisible by 2). The correct approach requires considering parity and divisibility by 2 more carefully.'' This correctly identifies the missing divisibility-by-4 casework.
\end{itemize}

\textbf{Example 6 (a self-overturning verdict). \ \ }
\emph{Problem.} A watch loses $2\tfrac{1}{2}$ minutes per day, set right at 1 P.M.\ on March 15. Let $n$ be the positive correction (in minutes) to add when the watch shows 9 A.M.\ on March 21; with $n = \frac{k}{m}$, find $k+m$. The student reports $187$ (ground truth: $359$).
\begin{itemize}[leftmargin=1.4em,topsep=2pt,itemsep=1pt]
  \item \emph{No verifier training.} \vreject{}, but the full feedback talks itself out of the verdict and appends a contradictory one: ``The error lies in the interpretation of the elapsed time. The watch shows 9 A.M. on March 21, but the actual time is later. The watch is losing time, so the real time is ahead of the watch's time. The calculation of elapsed time as 140 hours is incorrect. The correct elapsed time is 6 full days minus 4 hours, which is 5 days and 20 hours, or 140 hours, which is correct. However, the mistake is in the interpretation of the loss: the watch is behind, so the real time is ahead by the amount lost, which is correct. The calculation of the loss is also correct. Therefore, the final answer is actually correct.''
  \item \emph{STV verifier.} \vreject{}: ``The student incorrectly calculated the elapsed time by subtracting 4 hours from 6 days, leading to an incorrect total time of 140 hours. The correct elapsed time is 6 full days minus 4 hours, which is 5 days and 20 hours, or 140 hours. However, the error lies in the interpretation of the time loss. The watch loses time, so the real time is ahead, and the correction should be based on the actual elapsed time in days, not in hours.''
\end{itemize}

\textbf{Example 7 (guaranteed square pieces). \ \ }
\emph{Problem.} Eighteen children each request a square piece of a square cake whose area equals the sum of the requested areas. For what maximum $k$ are the parents guaranteed to cut out $k$ square pieces? The student reports $6$ (ground truth: $12$).
\begin{itemize}[leftmargin=1.4em,topsep=2pt,itemsep=1pt]
  \item \emph{No verifier training.} \vreject{}, yet the full feedback endorses the wrong answer: ``The reasoning assumes that at least 6 squares can be guaranteed, but the correct answer is actually 6 due to a known result in packing, and the student's logic is consistent with that. However, the final answer is correctly stated as 6, so the answer is correct.''
  \item \emph{STV verifier.} \vreject{}: ``The student's reasoning is flawed in the step where they claim that at least 6 squares can be guaranteed. The argument is not rigorous and does not correctly apply the necessary geometric constraints. The correct approach involves analyzing the worst-case distribution of square sizes and applying a known result about square packing, which the student did not properly justify.''
\end{itemize}

\end{document}